\documentclass[10pt,a4paper]{article}
\usepackage[utf8]{inputenc}
\usepackage[english]{babel}
\usepackage{amsmath}
\usepackage{amsfonts}
\usepackage{amssymb}
\usepackage{graphicx}
\usepackage{lmodern}
\usepackage{cite}
\usepackage{siunitx}
\usepackage{fixmath}

\renewcommand{\vec}[1]{\mathbf{#1}}
\newcommand{\mat}[1]{\mathrm{#1}}

\usepackage{tikz}
\usetikzlibrary{math}
\usetikzlibrary{positioning}

\usepackage{hyperref}
\hypersetup{colorlinks,citecolor=black,filecolor=black,linkcolor=black,urlcolor=black}

\author{Matti Pellikka\textsuperscript{1}, Valtteri Lahtinen\textsuperscript{2}}
\title{A Robust Method for Image Stitching}

\begin{document}

\maketitle

\noindent\textsuperscript{1}Grundium Ltd, Tampere, Finland \\ 
\textsuperscript{2}QCD Labs, QTF Centre of Excellence, Department of Applied Physics, Aalto University, 00076, Aalto, Finland

\begin{abstract}
We propose a novel method for large-scale image stitching that is robust against repetitive patterns and featureless regions in the imagery. In such cases, state-of-the-art image stitching methods easily produce image alignment artifacts, since they may produce false pairwise image registrations that are in conflict within the global connectivity graph. Our method augments the current methods by collecting all the plausible pairwise image registration candidates, among which globally consistent candidates are chosen. This enables the stitching process to determine the correct pairwise registrations by utilizing all the available information from the whole imagery, such as unambiguous registrations outside the repeating pattern and featureless regions. We formalize the method as a weighted multigraph whose nodes represent the individual image transformations from the composite image, and whose sets of multiple edges between two nodes represent all the plausible transformations between the pixel coordinates of the two images. The edge weights represent the plausibility of the transformations. The image transformations and the edge weights are solved from a non-linear minimization problem with linear constraints, for which a projection method is used. As an example, we apply the method in a large-scale scanning application where the transformations are primarily translations with only slight rotation and scaling component. Despite these simplifications, the state-of-the-art methods do not produce adequate results in such applications, since the image overlap is small, can be featureless or repetitive, and misalignment artifacts and their concealment are unacceptable.
\end{abstract}

\section{Introduction}
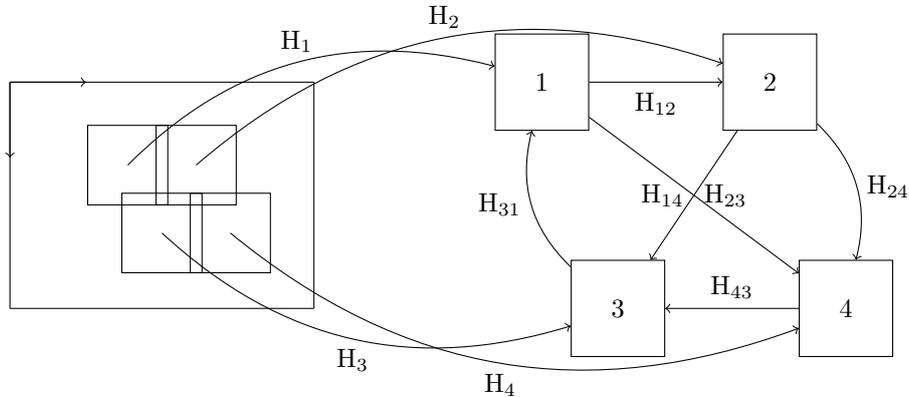
\begin{figure*}[ht]
\centering
\begin{tikzpicture}
\tikzmath{  \imsize = 15; \no=-0.4; \x1 = -2; \y1 = 2; \x2 = 2; \y2 = -1;}
\begin{scope}[node distance=0.9 and 0.45, inner sep = \imsize, on grid, every node/.style={rectangle}]
\node [draw] (I1) at (0,0) {}; 
\node [draw] (I2) [above left=of I1] {};
\node [draw] (I3) [above right=of I1] {}; 
\node [draw] (I4) [below right=of I3] {}; 
%
\node [draw] (J3) at (6,-1) {3} edge [bend left, <-] node [below=\no] {$\mat{H}_{3}$} (I1.center);
\node [draw] (J1) at (5,2) {1} edge [bend right, <-] node [above=\no] {$\mat{H}_{1}$} (I2.center);
\node [draw] (J2) at (8,2) {2} edge [bend right, <-] node [above=\no] {$\mat{H}_{2}$} (I3.center);
\node [draw] (J4) at (9,-1) {4} edge [bend left, <-] node [below=\no] {$\mat{H}_{4}$} (I4.center);
\draw (J1) edge [->] node [below=\no] {$\mat{H}_{12}$} (J2);
\draw (J2) edge [bend left, ->] node [right=\no] {$\mat{H}_{24}$} (J4);
\draw (J4) edge [->] node [above=\no] {$\mat{H}_{43}$} (J3);
\draw (J3) edge [bend left, ->] node [left=\no] {$\mat{H}_{31}$} (J1);
\draw (J1) edge [->] node [left=\no] {$\mat{H}_{14}$} (J4);
\draw (J2) edge [->] node [right=\no] {$\mat{H}_{23}$} (J3);

\end{scope}

\coordinate (A) at (\x1,\y1);
\coordinate (B) at (-1,\y1);
\coordinate (C) at (\x1,1);

\draw [->] (A) -- (B);
\draw [->] (A) -- (C);
\node [above] at (B) {};
\node [left] at (C) {};

\draw [-] (\x1,\y1) -- (\x1,\y2);
\draw [-] (\x1,\y1) -- (\x2,\y1);
\draw [-] (\x1,\y2) -- (\x2,\y2);
\draw [-] (\x2,\y1) -- (\x2,\y2);

\end{tikzpicture}
\caption{Transformations $\mat{H}_i$ from the composite image pixel coordinates to the individual overlapping images and transformations $\mat{H}_{ij}$ between image pairs. In the pairwise registration phase, the transformations $\mat{H}_{ij}$ are found, while $\mat{H}_i$ are found in the subsequent global alignment phase and are used to render the composite image using image warping.}
\label{fig:overview}
\end{figure*}

Stitching is a process of creating a seamless composite image from a set images that cover a desired scene or surface. The objective is to find a transformation of pixel coordinates from the composite image to each individual image in order to be able to render the corresponding region in the composite image, depicted in Figure~\ref{fig:overview}. In this paper, we introduce a novel, robust method for image stitching.

\subsection{Background}

The earliest methods and applications of image stitching originate from photogrammetry, where aerial photographs were manually registered using ground control points to produce composite images for aerial survey \cite{Wolf2000McGrawHill}. With digital image processing, methods for automatic image registration were developed to eliminate the manual labor of locating the ground control point patterns. Also, methods to register images without artificial control point patterns were developed, where either directly the image intensity patterns, such as for example intensity gradient \cite{10.5555/1623264.1623280} or intensity cross-correlation \cite{Lewis95fastnormalized}, or previously detected image-intensity-based feature points, such as for example SIFT \cite{10.5555/850924.851523}, SURF \cite{10.1016/j.cviu.2007.09.014}, or ORB \cite{10.1109/ICCV.2011.6126544}, were used to guide the registration. With such methods, coping with false matches becomes an intrinsic part of the image registration process, as the matched patterns can be subtle, misleading, missing entirely, or repetitive. The typical method is to sample a large body of feature point correspondence candidates and use a robust estimation method such as RANSAC \cite{fischler_bolles_1981} to filter out the false matches.

The majority of the current image stitching methods use projective transformations, i.e. homographies, as the motion model in the pairwise registration. This is justified when the camera motion is almost rotational only, or the imaged scene is effectively planar. Otherwise, the use of rigid homographies may result in parallax errors, as the foreground objects move with respect to the scene background between different views of the scene. Much of the recent research tries to solve this problem without a full 3D reconstruction of the scene by producing non-rigid pairwise registrations. For example, in \cite{journals/pami/ZaragozaCTBS14} local non-projective deviations are allowed. In  \cite{Jacob_2018_ECCV_Workshops}, conformal mappings with small non-conformal deviations are used. Local and interpolated homographies are determined in \cite{7298719, 10.1109/CVPR.2014.423}, and similarity transformations in \cite{8099772}.

The pairwise registration is not sufficient to produce a seamless composite image. If the images are tiled together one by one, small errors in the pairwise registrations accumulate, creating noticeable misalignment. Therefore, methods to solve for globally consistent registrations were developed, often called \emph{bundle adjustment} or \emph{global alignment}. A detailed description is provided in \cite{10.1007/3-540-44480-7_21}. In \cite{10.1007/s11263-006-0002-3}, a full process from feature detection to global alignment is described. In such methods, the image or camera locations are solved for by minimizing some error metric that utilizes the information from all the pairwise registrations, and small registration errors are likely to cancel out. 

\subsection{Motivation for a new method}
Despite the above advancements, the current methods and tools for image stitching are still susceptible to producing stitching artifacts, as possible false pairwise image registrations are still considered in the bundle adjustment phase and result in misaligned images. The state-of-the-art methods often try to conceal the misalignment artifacts in the image blending phase, producing a convincing-looking but not true-to-reality composite image. In the consumer mobile-camera panorama stitching applications, this is often tolerated. Also, in the case of non-rigid pairwise image registration, small scale fidelity is often achieved on the expense of large-scale deformation of the scene.

In our method, we improve the image alignment phase robustness in the challenging situations where the image overlap is small, and may contain repetitive or non-existent texture. We achieve this by extending the global alignment phase to have similar robustness characteristics as the pairwise registration phase. That is, we sample multiple plausible pairwise registrations, and filter out the false ones during the global alignment phase. Thus, both of the main phases of the image stitching process become tolerant to outliers, increasing the overall robustness. Instead of being a replacement, our method can be seen as an additional step to many current stitching applications to improve their robustness. 

We limit our scope to rigid transformations between the images and draw our examples from a scanning application. Here, the imaged scene is assumed to be planar and the motion model is almost pure translation, and the image connectivity is known beforehand. In this application however, the misalignment artifacts are not tolerated and their concealment is unacceptable, but the image overlap is made as small as possible and the number of images is large. Despite the problem setting is simplified in some aspects, the state-of-the-art methods may misalign the images due to including false pairwise registrations on regions of non-existent or repetitive texture, or by leaving them out completely. These false or missing pairwise alignments then easily affect the result of the global adjustment phase, producing a malformed composite image.

After the first two phases, pairwise registration and global alignment, \emph{image blending} can be considered as the third main phase of image stitching. In this phase, the seams arising from image exposure differences are faded out \cite{10.1145/245.247, 10.1145/882262.882269} and the seam path is selected to avoid ghosts arising from objects that change positions between the image captures \cite{991005, 1641060}. We do not discuss this phase further, but point out that these effects also decrease the pairwise image registration robustness. As our method is able to consider multiple pairwise registration candidates, it is able to improve the overall image alignment robustness. Otherwise it might get misguided by the false pairwise registrations due to moving objects.

Previous methods to improve stitching robustness in presence of repetitive patterns focus on improving the pairwise registration phase by trying to avoid false registrations altogether. Instead of repetitive point features, they often rely on large scale features \cite{6999708,5597189} or feature point clusters \cite{Ha2012ImageRB} that are no longer repeating. Such methods require a large overlapping region between the image pair. In scanning applications where the scanning speed should be as fast as possible, such methods are not preferable. Another approach is to use a robust estimation method, such as Huber loss \cite{huber:1964}, in the global alignment phase to be able to neglect false pairwise registrations \cite{10.1007/s11263-006-0002-3}. With such approaches however, false pairwise registrations still contribute to the loss. When the majority of the pairwise registrations are false, the minima of the loss function are not guaranteed to correspond to consistent cycles of transformations in the connectivity graph, while the minima in our method have this property.

\subsection{Our method: A walkthrough}

In this section, we give an informal overview of our method and its benefits. In section \ref{seq:method}, we describe the method in detail.

Typically, stitching starts with a pairwise registration phase of images that partially overlap \cite{10.1561/0600000009}. The registration is based on finding closely resembling feature point pairs in the overlapping regions to determine a transformation between the pixel coordinates of the image pair. A robust estimation method such as RANSAC \cite{fischler_bolles_1981} is typically used to find the transformation together with a set of true point correspondences, as the majority of the proposed point correspondences can be outliers. Each pairwise transformation constitutes an edge in a connectivity graph, whose nodes represent the unknown pixel coordinate transformations from the composite image to each image. Once all the image pairwise registrations are found, the nodal transformations from the composite image to each image are found by minimizing the transformation error between all the feature point correspondences in the connectivity graph. Finally, the transformations are used to render the composite image with a process called image warping, described in detail for example in \cite{10.5555/1941882}.

If the overlapping region of an image pair contains a repetitive texture or feature pattern, is devoid of any detail, or is a moving content, the above method easily produces an incorrect transformation and point correspondence set for the image pair. This is because a false and a true set of point correspondences might seem equally likely, in terms of feature point pair similarity and correspondence set size. The false correspondences cause visible stitching artifacts in the composite image, as the edge transformations in the connectivity graph contain conflicting or false information. For example, a cycle of transformations in the connectivity graph does not necessarily produce an identity transformation. 

Our image stitching method is robust against such repetitive patterns in the imagery or otherwise false pairwise registrations, given the connectivity graph contains at least one cycle. Instead of a simple connectivity graph we construct a connectivity multigraph, where we find all the plausible transformations and point correspondence sets between an image pair, and include each of them as an edge in the connectivity graph between the two nodes. A dummy edge is also included, for the possibility that none of the edge candidates is a true match. 

Once the multigraph has been constructed, we solve for the nodal transformations from the composite image pixel coordinates to each image pixel coordinates together with the edge weights that represent the edge plausibility. The method is formulated as a non-linear minimization problem with the constraint that for each set of edges between two nodes, the edge weights sum to one. The edges weights can thus be interpreted as the plausibility of the corresponding registration to be the correct one. We solve the equality constrained minimization problem using a projection method. Once the minimization problem has been solved, for each set of edges, the edge with the largest weight is regarded as a true registration. If the dummy edge weight is the largest, all the original edges are considered as false registrations.

Our method has the most benefit in stitching tasks where the image connectivity graph has plenty of cycles, for example in a scanning application. This is because for consistency, each cycle of transformations is required to produce an identity transformation which is enforced by the minimization procedure. If a cycle contains image regions outside the repeating or featureless regions where unambiguous pairwise registrations can be obtained, this method propagates the reliable registration information along the cycle to the ambiguous regions. On the contrary, in the cases where the connectivity graph is acyclic, our method cannot distinguish between true and false pairwise registrations.

\subsection{Structure of this paper}

In section \ref{seq:method}, we describe the proposed method in detail. First, the process of obtaining a multigraph of pairwise image registration candidates is provided. Second, we formalize and give a solution method to the problem of performing a global alignment of the images given the pairwise registration candidates. In section \ref{seq:experiments}, we give three examples for which the proposed method provides a clear improvement compared to the standard approaches. Each example showcases a different benefit of our method. Finally, in section \ref{seq:conclusion}, we draw conclusions.

\section{Proposed method} \label{seq:method}

Like the typical stitching process, our method has two phases. First, the pairwise registration phase finds all the plausible transformations between an image pair for all possible image pairs. Second, the global alignment phase finds an agreement between all the pairwise registrations, filtering out all the pairwise transformations that fit poorly to the rest of the connectivity graph.

While we give a general formulation of our method, we provide example steps and definitions that work well in our scanning application. However, the method should generalize for other stitching tasks, and we point out some alternative definitions that might be more suitable in other contexts.

\subsection{Multigraph construction}

Given a set of images, we construct a multigraph containing all the plausible transformations $\{ \mat{H}_{ij,k} \}_k$ from pixel coordinates of image $i$ to image $j$ pixel coordinates. For brevity, we consider the transformation $\mat{H}_{ij,k}$ to be perspective transformation, represented by a $3 \times 3$ matrix acting on homogeneous pixel coordinates. However, the method generalizes to continuously differentiable change of coordinates between the images, for example, by including a parametric lens distortion correction to the transformation, and also specializes to mere translations between the images. Typically, not all the image pairs need to be attempted to be registered with each other, as it is often known in advance which images overlap, for example from the approximate camera motion.

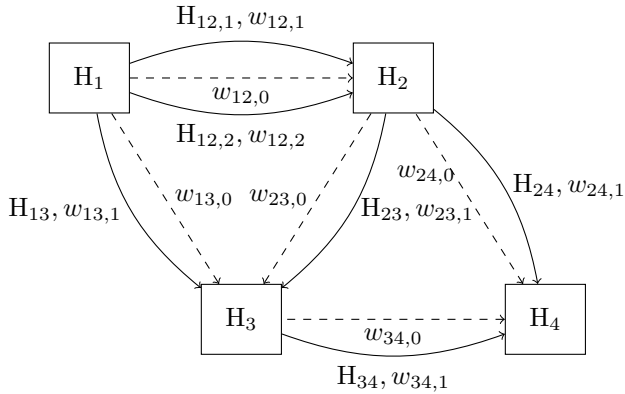
\begin{figure}[ht]
\centering
\begin{tikzpicture}
\tikzmath{\imsize = 9; \ba = 20; \no=-0.2;}

\begin{scope}[node distance=3.2 and 2.0, inner sep = \imsize, on grid, every node/.style={rectangle}]
\node [draw] (I3) at (0,0) {$\mat{H}_3$}; 
\node [draw] (I2) [above right=of I3] {$\mat{H}_2$} edge [bend left=\ba, ->] node [right=\no] {$\mat{H}_{23}, w_{23,1}$} (I3) edge [dashed, ->] node [left=\no] {$w_{23,0}$} (I3); 
\node [draw] (I1) [above left=of I3] {$\mat{H}_1$} edge [bend left=\ba, ->] node [above=\no] {$\mat{H}_{12,1}, w_{12,1}$} (I2) edge [bend right=\ba, ->] node [below=\no] {$\mat{H}_{12,2}, w_{12,2}$} (I2) edge [bend right=\ba, ->] node [left=\no] {$\mat{H}_{13}, w_{13,1}$} (I3) edge [dashed, ->] node [right=\no] {$w_{13,0}$} (I3) edge [dashed, ->] node [below=\no] {$w_{12,0}$} (I2); 
\node [draw] (I4) [below right=of I2] {$\mat{H}_4$} edge [dashed, <-] node [above left=\no] {$w_{24,0}$} (I2) edge [dashed, <-] node [below=\no] {$w_{34,0}$} (I3) edge [bend right=\ba, <-] node [right=\no] {$\mat{H}_{24}, w_{24,1}$} (I2) edge [bend left=\ba, <-] node [below=\no] {$\mat{H}_{34}, w_{34,1}$} (I3);

\end{scope}

\end{tikzpicture}
\caption{A multigraph with two transformation candidates for the edge $12$, and with one candidate for the other edges. The dashed lines represent the dummy edges.}
\label{fig:multigraph1}
\end{figure}

The transformations $\mat{H}_{ij,k}$ between the image pairs $ij$ are presented by the edges of the multigraph. The unknown edge weights $w_{ij,k}$ represent the plausibility of the transformation and the unknown transformations $\mat{H}_{i}$ from the composite image to each image are presented by the nodes of the multigraph. Finally, a dummy edge with an unknown weight $w_{ij,0}$ is added to each set of edges. A small example multigraph structure is represented in Figure \ref{fig:multigraph1}, where the edge $12$ has two plausible transformations $\mat{H}_{12,1}$ and $\mat{H}_{12,2}$ and the dashed lines represent the dummy edges.

Any method for finding the set $\{ \mat{H}_{ij,k} \}_k$ of plausible transformations between the image pairs can be used. For example, RANSAC \cite{fischler_bolles_1981} can be modified to produce multiple point correspondence consensus sets. Once a consensus set is found, its point correspondences are removed from the initial set of point correspondence candidates, and RANSAC is rerun with the remaining point correspondences to find another consensus set. In a scanning application, where the transformations are primarily approximately known translations, another method of obtaining multiple plausible transformations is to find all the translations corresponding to strong local maxima in the image cross-correlation \cite{Lewis95fastnormalized}. To illustrate the possibility of multiple plausible transformations between an image pair, we describe the cross-correlation method in detail. 

\subsubsection{Multiple transformation candidates with cross-correlation}

\begin{figure}[ht]
\centering
\begin{tikzpicture}
\tikzmath{ \ax = 2.5; \ay = 0.5; \bx = 2.6; \by=1; \imsize = 10; }
\node [draw, rectangle, inner sep = 35] at (1.5,1.5) {$I_i$};
\node [draw, rectangle, inner sep = 35] at (5.5,0.5) {$I_j$};

\coordinate (A) at (2.5,0.5);
\coordinate (B) at (2.6,1.5);
\coordinate (a) at (4.5,0.2);
\coordinate (b) at (4.6,1.2);
\coordinate (a2) at (4.5,0.1);
\coordinate (b2) at (4.6,1.1);
\draw  (A) edge [->] node[above] {$\Delta \vec{x}_0$} (a);
\draw  (B) edge [->] node[above] {$\Delta \vec{x}_0$} (b);

\node [above] at (A) {$\vec{x}_2$};  
\node [above] at (B) {$\vec{x}_1$};  

\node [draw, rectangle, inner sep = 7] at (a) {};  
\node [draw, rectangle, inner sep = 7] at (b) {}; 

\draw[fill] (A) circle [radius=0.025];
\draw[fill] (B) circle [radius=0.025];
\draw[fill] (a2) circle [radius=0.025];
\draw[fill] (b2) circle [radius=0.025];

\end{tikzpicture}
\caption{Cross-correlation computation with two feature points $\vec{x}_p$.}
\label{fig:correlation}
\end{figure}
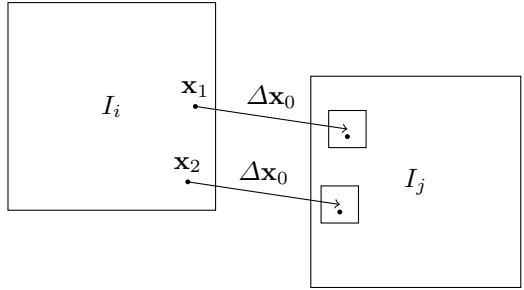

For partially overlapping image pair, the image cross-correlation computation can be accelerated by first finding a set of distinctive features points from the first image, for example by using the Harris corner detector \cite{Harris88acombined, Shi94goodfeatures}. Given such a set of distinctive feature points $\{\vec{x}_p\}$ in the first image of an image pair, the local maxima of the Pearson correlation coefficient \cite{royal1895proceedings, Lewis95fastnormalized} 
\begin{align}
\rho_{ij}(\Delta\vec{x}) &= \frac{1}{N}\sum_{p=1}^N \frac{\sum \hat{I}_i(\vec{x}_p) \hat{I}_j(\vec{x}'_p)}{\sqrt{\sum \hat{I}_i(\vec{x}_p)^2 \sum \hat{I}_j(\vec{x'}_p))^2}},
\end{align}
can be used as the correlation metric. Here $p$ is the indexing of the feature points, and $\hat{I}_i(\vec{x}) = I_i(\vec{x}) - \bar{I}_i(\vec{x})$ holds, where $I_i(\vec{x})$ and $\bar{I}_i(\vec{x})$ denote the image $i$ pixel value at $\vec{x}$ and the mean value around a neighborhood of $\vec{x}$, respectively, and $\vec{x}'_p = \vec{x}_p + \Delta \vec{x}_0 + \Delta \vec{x}$ holds, where $\Delta \vec{x}$ is the deviation from the initial translation estimate $\Delta \vec{x}_0$. The sums and means are taken on a square window around each feature point, see Figure \ref{fig:correlation}. 

\begin{figure}[ht]
\centering
\includegraphics[width=0.24\textwidth]{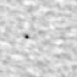}
\includegraphics[width=0.24\textwidth]{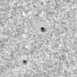}
\includegraphics[width=0.24\textwidth]{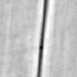}
\includegraphics[width=0.24\textwidth]{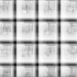}
\caption{Examples of image cross-correlations $\rho_{ij}(\Delta \vec{x})$, darker shade denotes stronger correlation. From left to right, top to bottom: a single strong correlation maximum, two maxima, a line of well correlated translations, correlation produced by a grid pattern in the images.}
\label{fig:correlations}
\end{figure}

The local maxima $\{\Delta\vec{x}_k\}$ of the function $\rho_{ij}(\Delta \vec{x})$ over some threshold are found and the corresponding sets $\{ \vec{x}_p \leftrightarrow \vec{x}_p + \Delta \vec{x}_0 + \Delta \vec{x}_k \}$ of the feature points $\vec{x}_p$ for each maximum $k$ are selected as candidate matches between the images $I_i$ and $I_j$. See Figure \ref{fig:correlations} for examples of cross-correlation scores $\rho_{ij}(\Delta \vec{x})$ that can be encountered. 

\subsection{Multigraph alignment}

Once the multigraph has been constructed, we solve for the edge weights $w_{ij,k}$ and the nodal transformations $\mat{H}_{i}$ simultaneously from the following non-linear minimization problem with linear constraints:
\begin{align}
& \min_{\vec{h}, \vec{w}}  f(\vec{h}, \vec{w}) = \nonumber \\ 
& \min_{\{\mat{H}\}, \{w\}} \sum_{i < j} \Big(w_{ij,0}^2\tau^2 + \sum_{k=1}^{N_e} w_{ij,k}^2 \| \mat{H}_{ij,k}\mat{H}_i -  \mat{H}_j \|^2 \Big), \label{eq:loss} \\  
& \textrm{subject to} \quad \sum_{k=0}^{N_e} w_{ij,k} = 1 \quad \forall \, ij, \, i<j,
\label{eq:constraint}
\end{align}
where the parameters vectors $\vec{h}$ and $\vec{w}$ of the loss function $f(\vec{h}, \vec{w})$ represent the unknown parameters of the nodal transformations $\mat{H}_{i}$ and the unknown edge weights $w_{ij,k}$, respectively, and $\tau$ is a fixed threshold. The transformation candidates $\mat{H}_{ij,k}$ are given by the pairwise registration phase. The norm $\| \cdot \|$ between the transformations is chosen be suitable for the type of transformations $\mat{H}_i$. For example, in the case of pure translations, we choose the Euclidean 2-norm $\| \Delta \vec{x}_{ij,k} + \Delta \vec{x}_i - \Delta \vec{x}_j\|$. For perspective transformations, one could use for example a homographic 2-norm \cite{JE2015185}, and for more general transformations the $L^2$-norm.

The rationale of this minimization problem is to find from each set $\{ \mat{H}_{ij,k} \}_{k=1}^{N_e}$  of edge transformations between two nodes $i$ and $j$ the transformation $\mat{H}_{ij,k}$ that is the best fit with respect to the rest of the graph. The plausibility of the transformation $\mat{H}_{ij,k}$ is given by the weight $w_{ij,k}$. The possibility that none of the given transformations fits well with the rest of the graph is expressed by the weight $w_{ij,0}$ of the dummy edge. The reference measure for this incompatibility is the threshold $\tau$. In other words, the cycle inconsistency for a cycle of $n$ edges is at most $n \tau$. The weight constraint \eqref{eq:constraint} is needed to avoid the trivial minima at $\vec{w} = 0$.

The minima of the loss function in \eqref{eq:loss} are guaranteed to be such that the cycles of transformations are consistent within the connectivity graph, i.e. the norm of a composite transformation along a cycle is at most $n\tau$, where $n$ is the cycle length. If the norm was larger, the loss function would obtain a smaller value if a dummy edge weight $w_{ij,0}$ was increased and the corresponding weights $w_{ij,k}$, $k > 0$ were decreased. In other words, the method will break any cycles that are in conflict with the cycle consistency. Therefore, the threshold value $\tau$ should not be set too low for the method to break legitimate cycles. That is, a suitable threshold value should adapt to the estimation uncertainty of the pairwise transformations $\mat{H}_{ij,k}$. In addition to the feature point localization accuracy, the uncertainty might depend for example on unaccounted lens distortion or scale differences.

\subsubsection{Constrained minimization}
The constraint equation \eqref{eq:constraint} can be expressed as a matrix equation $\mat{J}\vec{w} = \vec{1}$, where the matrix $\mat{J}$ consists of ones and zeros and $\vec{1}$ is a vector of ones. For example, for the graph of Figure \ref{fig:multigraph1}
\begin{align}
\setcounter{MaxMatrixCols}{20}
\mat{J}\vec{w} = \begin{bmatrix}
1 & 1 & 1 & 0 & 0 & 0 & 0 & 0 & 0 & 0 & 0\\
0 & 0 & 0 & 1 & 1 & 0 & 0 & 0 & 0 & 0 & 0\\
0 & 0 & 0 & 0 & 0 & 1 & 1 & 0 & 0 & 0 & 0\\
0 & 0 & 0 & 0 & 0 & 0 & 0 & 1 & 1 & 0 & 0\\
0 & 0 & 0 & 0 & 0 & 0 & 0 & 0 & 0 & 1 & 1
\end{bmatrix}
\begin{bmatrix}
\vec{w}_{12}\\
\vec{w}_{13}\\
\vec{w}_{23}\\
\vec{w}_{24}\\
\vec{w}_{34}
\end{bmatrix}
= 
\begin{bmatrix}
1 \\
1 \\
1 \\
1 \\
1
\end{bmatrix}
\end{align}
holds. It follows that in order for an iterative method to minimize the expression \eqref{eq:loss} subject to the constraint \eqref{eq:constraint}, the constrained update $\Delta\vec{w}$ must satisfy
\begin{align}
\mat{J}\Delta\vec{w} = \mat{J}\vec{w}_{n+1}-\mat{J}\vec{w}_n = \vec{1} - \vec{1} = \vec{0}.
\end{align}
That is, the update vector $\Delta\vec{w}$ must belong to the null-space of $\mat{J}$ to preserve the constraint on $\vec{w}$. Let the vectors $\{\vec{z}_i\}$ form the basis of this null-space. Given an unconstrained update $\Delta \vec{w}_u$, the constrained update $\Delta\vec{w}$ can be expressed as a linear combination
\begin{align}\label{eq:nullexp}
\Delta\vec{w} = \sum_i a_i \vec{z}_i := \sum_i (\Delta \vec{w}_u^\mat{T} \vec{z}_i) \vec{z}_i,
\end{align}
where the coefficients $a_i := \Delta \vec{w}_u^\mat{T} \vec{z}_i $ are the projections of the unconstrained updates $\Delta \vec{w}_u$ on the basis vectors $\vec{z}_i$ of the null space. Denote by $\mat{Z}$ the matrix whose columns are the basis vectors $\vec{z}_i$. Then, the equation \eqref{eq:nullexp} can be expressed in the matrix form:
\begin{align}
\Delta\vec{w} = \mat{Z}\vec{a} := \mat{ZZ^T}\Delta\vec{w}_u.
\end{align}
That is, the matrix $\mat{ZZ^T}$ projects the unconstrained update $\Delta \vec{w}_u$ to the null-space of $\mat{J}$. 

As a result, in any iterative minimization method, an unconstrained update rule $\vec{w}_{n+1} = \vec{w}_n + \Delta \vec{w}_u$ can be modified to satisfy the linear constraint \eqref{eq:constraint} by replacing the unconstrained update $\Delta \vec{w}_u$ with its projection $\mat{ZZ^T}\Delta\vec{w}_u$. This guarantees that the constraint \eqref{eq:constraint} is satisfied at each iteration if the initial guess satisfies it. 

For example, in the gradient descent method the update rule would be
\begin{align}
\vec{h}_{n+1} &= \vec{h}_{n} - \gamma_\vec{h} \nabla_{\vec{h}}f, \\
\vec{w}_{n+1} &= \vec{w}_{n} - \gamma_\vec{w} \mat{Z Z^T}\nabla_{\vec{w}}f, 
\end{align}
where $\nabla_{\vec{h}}f$ and $\nabla_{\vec{w}}f$ denote the gradient vectors of the loss function in \eqref{eq:loss} with respect to $\vec{h}$ and $\vec{w}$, respectively, and $\gamma_\vec{h}$ and $\gamma_\vec{w}$ are the step sizes. Here, the matrix $\mat{Z Z^T}$ projects the gradient vector $\nabla_{\vec{w}}f$ to the null-space of $\mat{J}$.

As the result of the minimization, one obtains the nodal transformations $\mat{H}_{i}$ that can be used as the image warp transformations to create the stitched composite image. 

\subsubsection{Faster convergence}
For faster convergence, Levenberg-Marquard method \cite{Levenberg1944, marquardt:1963} together with the conjugate gradient method can be used as follows. The second order Taylor expansion around a point $(\vec{h}_0, \vec{w}_0)$ of the loss function in the equation \eqref{eq:loss} is
\begin{align}
f(\vec{h}, \vec{w}) =& f(\vec{h}_0, \vec{w}_0) + 
\begin{bmatrix}
\nabla_{\vec{h}}f^\mat{T} &
\nabla_{\vec{w}}f^\mat{T}
\end{bmatrix}
\begin{bmatrix}
\Delta\vec{h}  \\
\Delta\vec{w}
\end{bmatrix} 
\label{eq:taylor} \\+& \frac{1}{2}
\begin{bmatrix}
\Delta\vec{h}^\mat{T} &
\Delta\vec{w}^\mat{T} 
\end{bmatrix}
\begin{bmatrix}
\nabla^2_{\vec{h}\vec{h}} f & \nabla^2_{\vec{h}\vec{w}} f \\
\nabla^2_{\vec{w}\vec{h}} f & \nabla^2_{\vec{w}\vec{w}} f 
\end{bmatrix}
\begin{bmatrix}
\Delta\vec{h}  \\
\Delta\vec{w} 
\end{bmatrix}, \nonumber
\end{align}
where $\nabla^2_{\cdot\cdot}f$ denote the Hessian matrix blocks of $f(\vec{h}, \vec{w})$ evaluated at $(\vec{h}_0, \vec{w}_0)$. To find a step $(\Delta \vec{h}, \Delta \vec{w})$ to minimize the Taylor expansion of $f(\vec{h}, \vec{w})$, the equation \eqref{eq:taylor} is differentiated with respect to $\Delta \vec{h}$ and $\Delta \vec{w}$, and the result is set to zero. This results in the equation
\begin{align}
\begin{bmatrix}
\nabla^2_{\vec{h}\vec{h}} f & \nabla^2_{\vec{h}\vec{w}} f \\
\nabla^2_{\vec{w}\vec{h}} f & \nabla^2_{\vec{w}\vec{w}} f 
\end{bmatrix}
\begin{bmatrix}
\Delta\vec{h}  \\
\Delta\vec{w}
\end{bmatrix} = 
\begin{bmatrix}
-\nabla_{\vec{h}}f \\ 
-\nabla_{\vec{w}}f
\end{bmatrix}.
\end{align}
Setting $\Delta \vec{w} = \mat{Z}\vec{a}$ to satisfy the constraint \eqref{eq:constraint}, and left-multiplying the second row with $\mat{Z}^\mat{T}$ results in
\begin{align}
\begin{bmatrix}
\nabla^2_{\vec{h}\vec{h}} f & \nabla^2_{\vec{h}\vec{w}} f \mat{Z} \\
\mat{Z}^\mat{T} \nabla^2_{\vec{w}\vec{h}} f & \mat{Z}^\mat{T} \nabla^2_{\vec{w}\vec{w}} f \mat{Z}
\end{bmatrix}
\begin{bmatrix}
\Delta\vec{h}  \\
\vec{a}
\end{bmatrix} = 
\begin{bmatrix}
-\nabla_{\vec{h}}f \\ 
-\mat{Z}^\mat{T} \nabla_{\vec{w}}f
\end{bmatrix}
\label{eq:taylor2}
\end{align}
from whose solution $\vec{a}$, the constraint-preserving step $\Delta \vec{w}$ can be recovered. The matrix of the equation \eqref{eq:taylor2} is usually very sparse, as typically one image overlaps with only a handful of other images to produce non-zero entries.

The matrix of the equation \eqref{eq:taylor2} may be indefinite, in which case step obtained from the solution of the system might not reduce the value of the loss function $f(\vec{h}, \vec{w})$. Downhill steps can be obtained by replacing the above equation with 
\begin{align}
\Bigg(
\begin{bmatrix} 
\nabla^2_{\vec{h}\vec{h}} f & \nabla^2_{\vec{h}\vec{w}} f \mat{Z} \\
\mat{Z}^\mat{T} \nabla^2_{\vec{w}\vec{h}} f & \mat{Z}^\mat{T} \nabla^2_{\vec{w}\vec{w}} f \mat{Z}
\end{bmatrix}
+ \lambda \mat{I} \Bigg)
\begin{bmatrix}
\Delta\vec{h}  \\
\vec{a}
\end{bmatrix} = 
\begin{bmatrix}
-\nabla_{\vec{h}}f \\ 
-\mat{Z}^\mat{T} \nabla_{\vec{w}}f
\end{bmatrix}, \label{eq:lm}
\end{align}
where $\mat{I}$ is the identity matrix, and $\lambda > 0$ is chosen to be such that the resulting matrix is positive definite and a downhill step is obtained. In practice, we choose an initial value for $\lambda$, and whenever a downhill step is obtained the value of $\lambda$ is reduced a little. If a non-positive definite matrix or an uphill step is encountered, the value of $\lambda$ is increased. Large values of $\lambda$ correspond to gradient descent steps. 

To further accelerate the convergence, we solve the sparse positive definite matrix equation \eqref{eq:lm} only approximately by taking just a few steps of the conjugate gradient method, i.e. stopping early. Our experiments show that the simple diagonal Jacobi preconditioner for the conjugate gradient method is efficient, as it does not reduce the sparsity of the system. While other preconditioners might speed up the convergence and reduce the number of steps, the loss of sparsity would increase the computational cost of each step.

\subsubsection{Global alignment of a pruned simple graph}
\begin{figure}[ht]
\centering
\begin{tikzpicture}
\tikzmath{\imsize = 10; \ba = 20; \no=-0.2;}

\begin{scope}[node distance=2.9 and 2.0, inner sep = \imsize, on grid, every node/.style={rectangle}]
\node [draw] (I3) at (0,0) {$\mat{H}_3$}; 
\node [draw] (I2) [above right=of I3] {$\mat{H}_2$} edge [dashed, ->] node [left=\no] {$w_{23,0}$} (I3); 
\node [draw] (I1) [above left=of I3] {$\mat{H}_1$} edge [bend left=\ba, ->] node [above=\no] {$\mat{H}_{12,1}, w_{12,1}$} (I2)  edge [bend right=\ba, ->] node [left=\no] {$\mat{H}_{13}, w_{13,1}$} (I3); 
\node [draw] (I4) [below right=of I2] {$\mat{H}_4$} edge [bend right=\ba, <-] node [right=\no] {$\mat{H}_{24}, w_{24,1}$} (I2) edge [bend left=\ba, <-] node [below=\no] {$\mat{H}_{34}, w_{34,1}$} (I3);

\end{scope}

\end{tikzpicture}
\caption{An example simple graph pruned from the multigraph of the Figure \ref{fig:multigraph1}. Transformation $\mat{H}_{12,1}$ was selected for the edge 12, and the candidate transformation $\mat{H}_{23}$ for the edge 23 was not compatible with the rest of the graph.}
\label{fig:simplegraph1}
\end{figure}
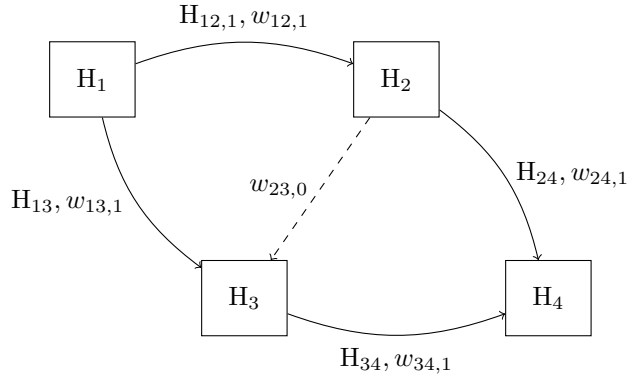

Alternatively, one can use the results of the previous minimization problem to prune the multigraph into a simple weighted graph for which the following typical global alignment procedure is applied. The obtained weights $w_{ij,k}$ are used as weighting factors associated with the point correspondence sets of the transformations $\mat{H}_{ij,k}$, so that only the point correspondence set with the largest weight $w_{ij} = \max_k \{w_{ij,k} \}$ is used in the global alignment. If the dummy edge gets the largest weight, all the associated point correspondence sets are discarded. That is, the edge is pruned altogether from the connectivity graph. See Figure \ref{fig:simplegraph1} for an example of a pruned graph.

Specifically, one solves the linear unconstrained minimization problem
\begin{align}
\min_{ \{\mat{H}_i\}} \sum_i \sum_j \sum_n w_{ij}^2 \| \mat{H}_i \vec{x}_{ij}^n - \mat{H}_j \vec{x}_{ji}^n \|^2,
\label{eq:pointalign} 
\end{align} 
where $w_{ij}$ is the largest weight for the edge $ij$, and the set $\{ \vec{x}_{ij}^n \leftrightarrow \vec{x}_{ji}^n \}$ is the set of point correspondences with the largest weight $w_{ij}$ for the edge $ij$.

The problem \eqref{eq:pointalign} can be rearranged as 
\begin{align}
\min_\vec{h} \| \mat{A}\vec{h} \|^2, 
\end{align}
where $\vec{h}$ is a stacked column vector of the parameters of the transformations $\mat{H}_i$. By construction, the null space of $\mat{A}$ has the dimension of the number of parameters in a transformation  $\mat{H}_i$. That is, one transformation needs to be fixed as the reference for the composite image coordinate system. We can divide the expression $\mat{A}\vec{h}$ into blocks
\begin{align}
\mat{A}\vec{h} = 
\begin{bmatrix}
\mat{A}_{11} & \mat{A}_{12} \\
\mat{A}_{21} & \mat{A}_{22} 
\end{bmatrix}
\begin{bmatrix}
\vec{h}_{1} \\
\vec{h}_{2} 
\end{bmatrix},
\end{align}
where $\vec{h}_{1}$ corresponding to the transformation $\mat{H}_1$ is chosen to be fixed. Then, the minimization problem transforms to 
\begin{align}
\min_{\vec{h}_2} \| \mat{A}_{21}\vec{h}_1 + \mat{A}_{22}\vec{h}_2  \|^2,
\end{align}
which has a unique solution given by 
\begin{align}
\vec{h}_2 = (\mat{A}_{22}^\mat{T}\mat{A}_{22})^{-1}\mat{A}_{22}^\mat{T}\mat{A}_{21}\vec{h}_1.
\end{align}

\section{Experiments} \label{seq:experiments}

In this section we present three example cases. The first case is often encountered in practice with digital microscope scanners: a tissue sample that contains some empty areas where pairwise matching produces ambiguous transformation candidates. The second one is an extreme case where almost all image pairs produce multiple transformation candidates: a microscope calibration slide with a repetitive pattern. The third case contains a pattern of tissue regions with void regions in between, which makes the determination of their placement challenging. For all the examples, the used input dataset for the stitching algorithm is available at \url{https://archive.org/details/stitching_202101}(accessed: May 3, 2021).

\subsection{Tissue sample with void regions}

\begin{figure}[ht]
\centering
\fbox{\includegraphics[width=0.6\textwidth]{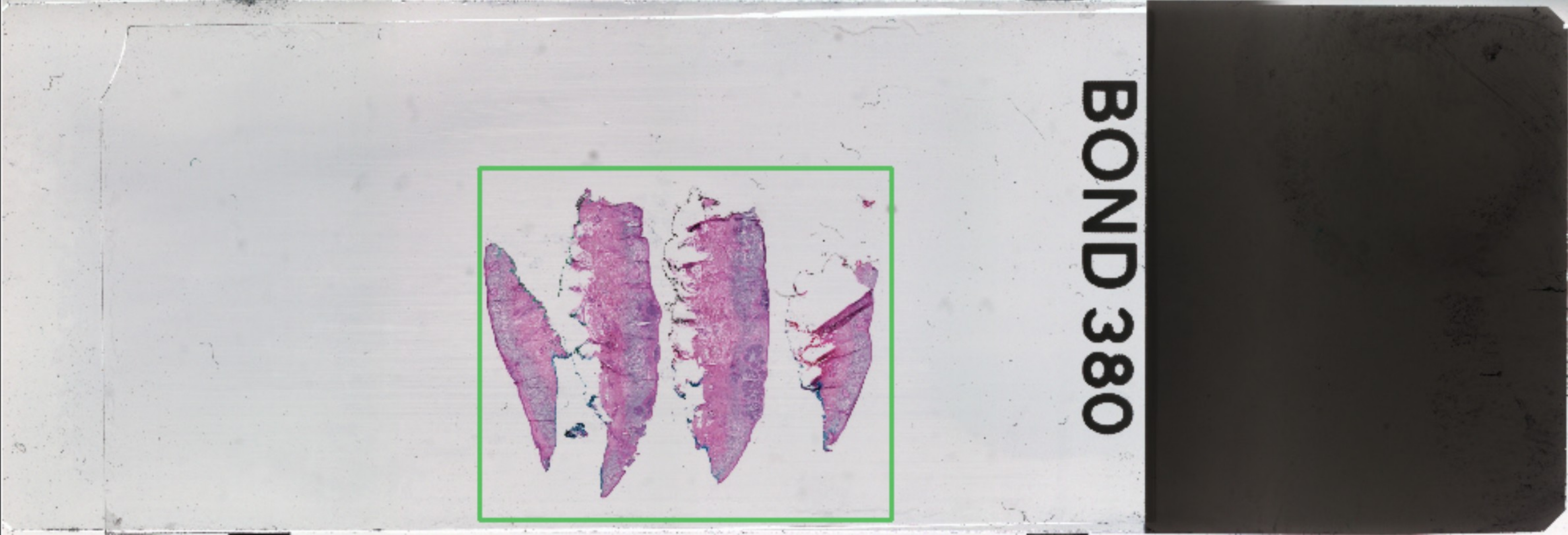}}
\caption{A typical tissue sample slide in pathology, an overview image. The green rectangle denotes the scanned region which contains void regions between the tissue slices. On a 20x digital microscope, the composite image of the 21 mm x 16 mm scan region constitutes about 1.3 gigapixels and is stitched from 348 fields of view.}
\label{fig:tissue}
\end{figure}

In a microscope scanning application in pathology, the sample specimen often consists of multiple slices of tissue with void regions in between, see Figure \ref{fig:tissue}. Within the void region, there might be very little or no details in the overlapping area of the images in the pairwise image registration, while within the tissue region the image registration works very reliably with the abundance of unique details. 

Our method increases the robustness of stitching with such sample as the method is able to provide multiple transformation candidates in the empty ambiguous region and possibly rule them all out by exploiting the information from the unambiguous transformations obtained from the connected tissue regions.

\begin{figure}[ht]
\centering
\includegraphics[width=0.9\textwidth]{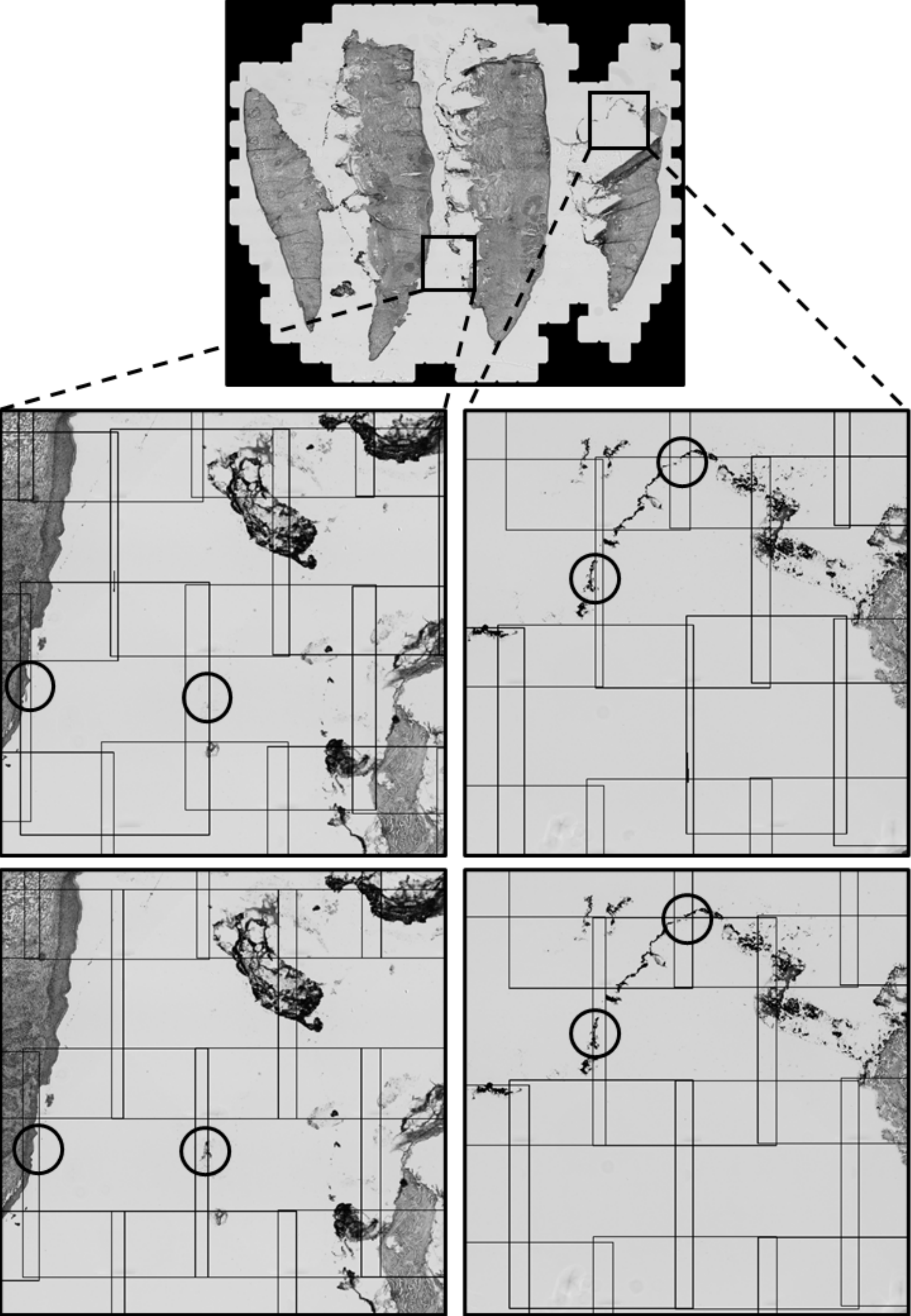}
\caption{On the first row, the transformation corresponding to the strongest cross-correlation is selected, while on the second row, our method with multiple transformation candidates together with the dummy candidate are considered. The black rectangles denote the image registrations after the global alignment. The black circles denote missing or malformed features in the first row, due to registration errors. Notice how a single false pairwise transformation also corrupts the nearby image locations in the global alignment, ill-forming the large-scale geometry.}
\label{fig:tissue1st}
\end{figure}

In Figure \ref{fig:tissue1st} we compare image patches where the pairwise transformations using the strongest cross\--cor\-re\-lation fails, whereas our method, where multiple candidate pairwise transformations are considered, succeeds. The resulting connectivity graph had 1472 edges, for which the dummy edge was selected for 24 edges, and for 2 out of 12 edges, transformation other than the transformation corresponding to the strongest cross-correlation was selected. 

\begin{figure*}[ht]
\centering
\fbox{\includegraphics[width=0.45\textwidth]{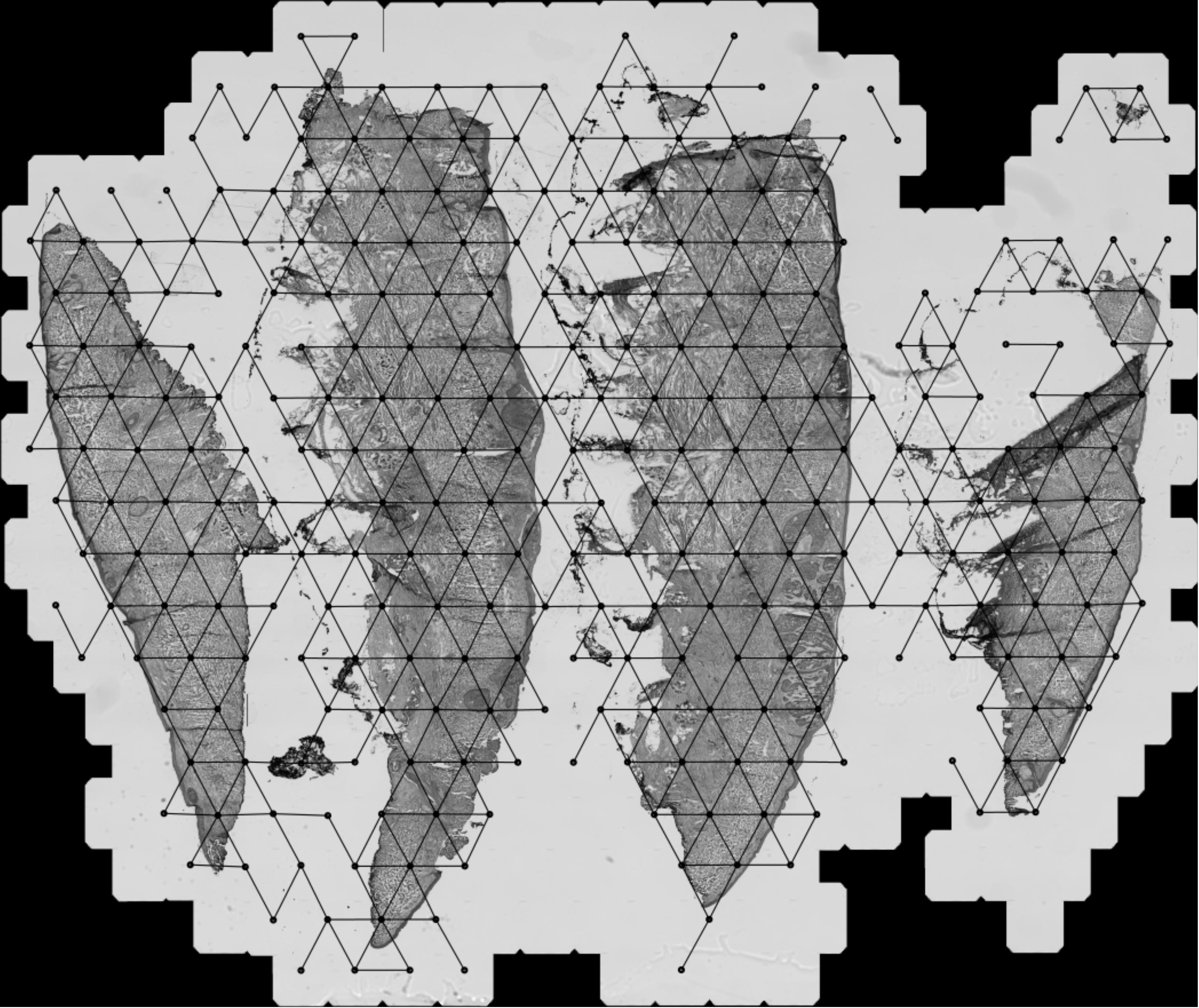}}
\fbox{\includegraphics[width=0.45\textwidth]{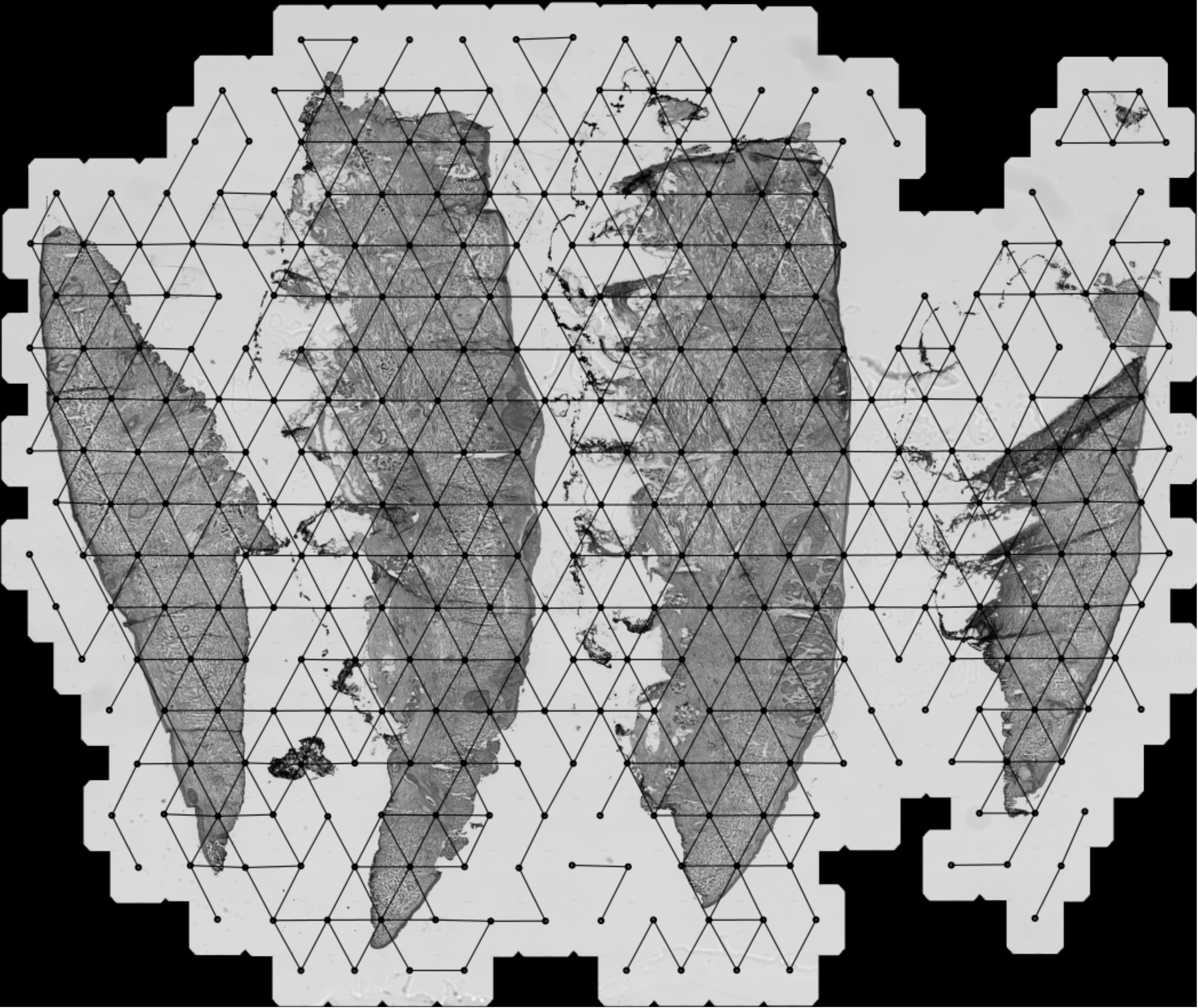}}
\caption{On the left is the final connectivity graph with the standard method, on the right with our method. Our method is able to produce more complete connectivity graph without introducing false pairwise matches, as it considers multiple pairwise transformation candidates and filters out the false ones. In contrast, the standard method is typically tuned to leave out uncertain pairwise registrations from the connectivity graph.}
\label{fig:tissue1stconn}
\end{figure*}

In Figure \ref{fig:tissue1stconn} we compare the connectivity graphs of our multigraph alignment method to the current standard method \cite{10.1007/s11263-006-0002-3} that uses ORB feature detector and RANSAC to find at most one transformation between the image pair. To avoid false pairwise registrations, the standard method leaves many of the pairwise transformations out from the connectivity graph. In the worst-case scenario, this could split the graph in multiple connected components, or leave a long detour path between neighboring images, causing a visible seam between the images.

\subsection{Calibration slide with a repetitive pattern}

\begin{figure}[ht]
\centering
\fbox{\includegraphics[width=0.5\textwidth]{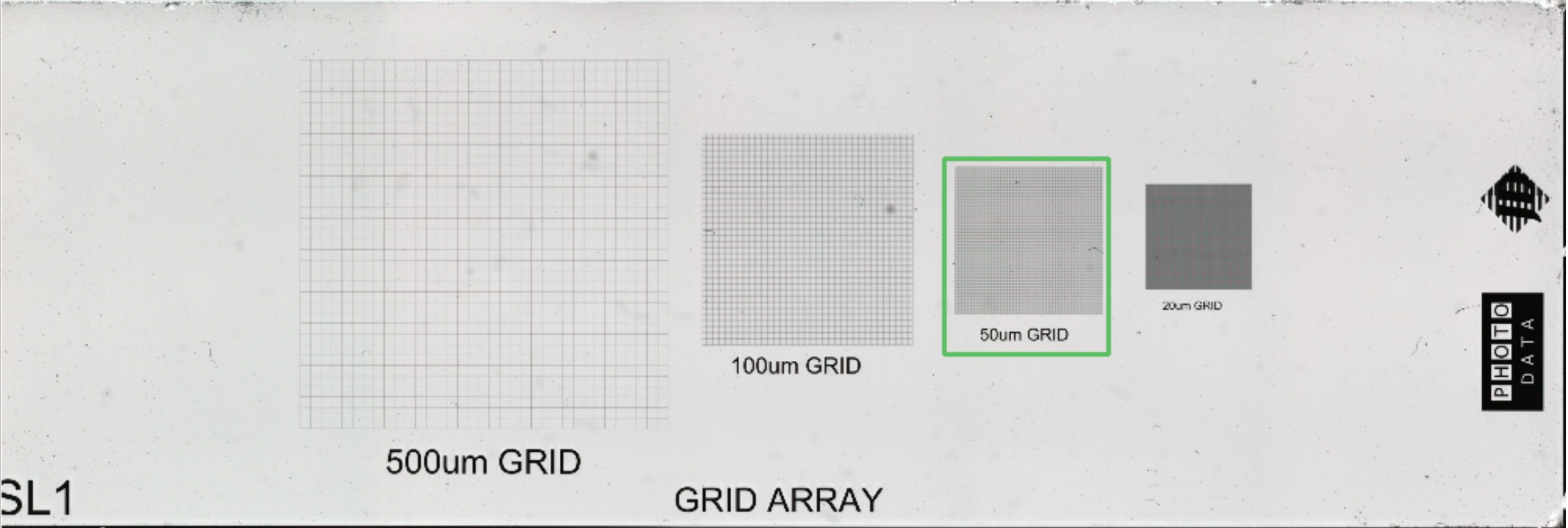}}
\caption{A calibration grid slide for microscope. Within the grid regions multiple equally likely transformations can be obtained from the pairwise image registration. The green rectangle denotes our example scanning area.}
\label{fig:grid}
\end{figure}

In Figure \ref{fig:grid}, we consider a microscope calibration grid slide where one cannot find unique pairwise transformations within the grid area. In this case, the pairwise image cross-correlation will produce multiple local maxima, as in the rightmost image in Figure \ref{fig:correlations}. Such a case is rarely encountered in practice, but some special sample slides can contain similar repeating marker patterns which are difficult to stitch. 

\begin{figure*}[ht]
\centering
\fbox{\includegraphics[width=0.45\textwidth]{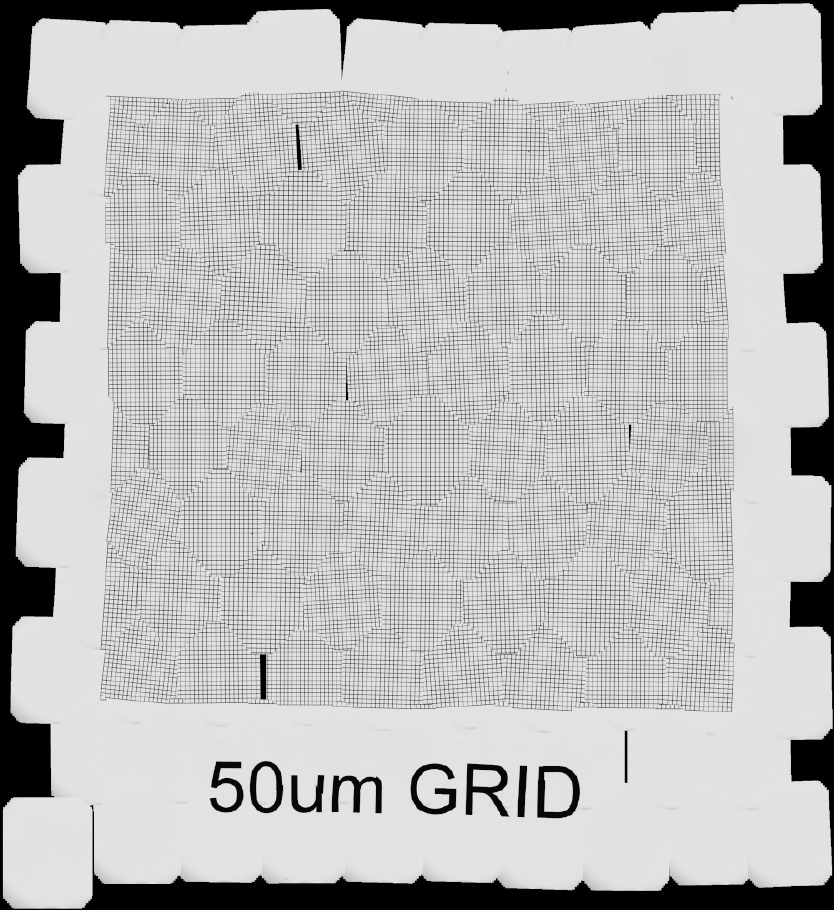}}
\fbox{\includegraphics[width=0.45\textwidth]{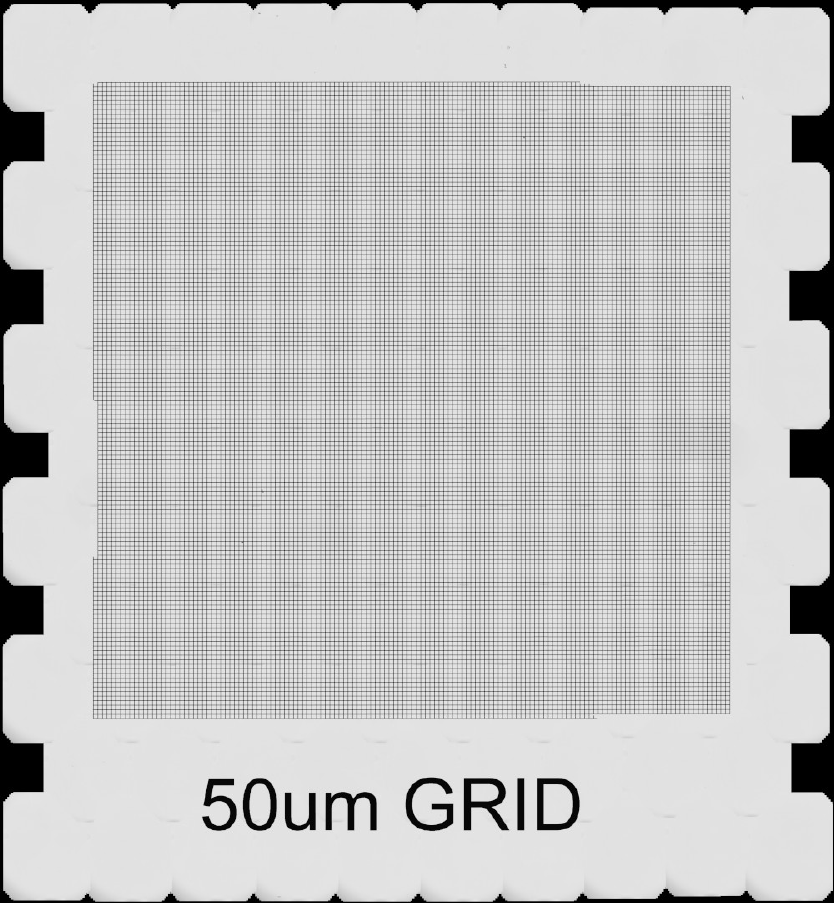}}
\caption{On the left, the transformation corresponding to the strongest cross-correlation is selected, while on the right, our method with multiple transformation candidates together with the dummy candidate are considered. Notice how the grid pattern is well aligned with our method, but the edges reveal that there exist false image registrations. The composite image is stitched from 105 fields of view.}
\label{fig:grid1st}
\end{figure*}

While our method produces the wrong result, the result appears convincing in the middle of the grid where even a human cannot distinguish the correct pairwise transformations from the false ones. Compare our result to the one obtained using the pairwise transformations corresponding to the strongest cross-correlations in Figure \ref{fig:grid1st}. The resulting connectivity graph had 486 edges, for which the dummy edge was selected for 12 edges, and for 262 out of 314 edges, transformation other than the transformation corresponding to the strongest cross-correlation was selected. 

\subsection{Sparse tissue pattern}

\begin{figure}[hb]
\centering
\fbox{\includegraphics[width=0.6\textwidth]{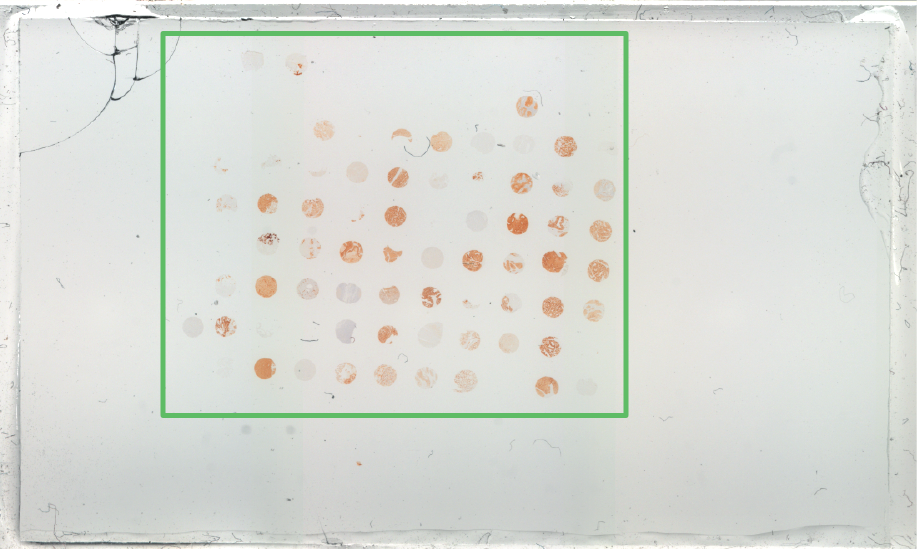}}
\caption{A microscope sample slide with a sparse pattern of tissue regions. The tissue regions are difficult to register to each other due to large void regions between them. The green rectangle denotes our example scanning area. On a 20x digital microscope, the composite image of the $23 \textrm{ mm} \times 19 \textrm{ mm}$ scan region constitutes about 1.7 gigapixels and is stitched from 492 fields of view.}
\label{fig:dots}
\end{figure}

In the example of Figure \ref{fig:dots}, a sparse pattern of tissue samples is placed on the sample slide so that grid spacing is slightly larger than a single field of view of a microscope. Therefore, no clear pairwise image registrations can be obtained across the void regions between the tissue regions.

\begin{figure*}[ht]
\centering
\fbox{\includegraphics[width=0.45\textwidth]{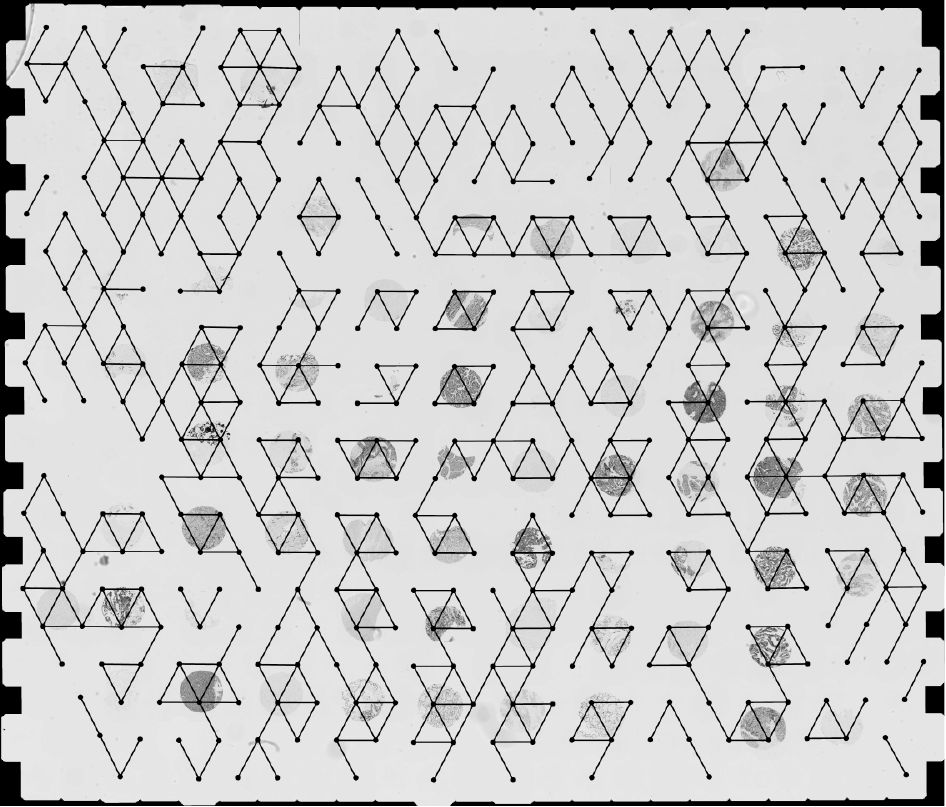}}
\fbox{\includegraphics[width=0.45\textwidth]{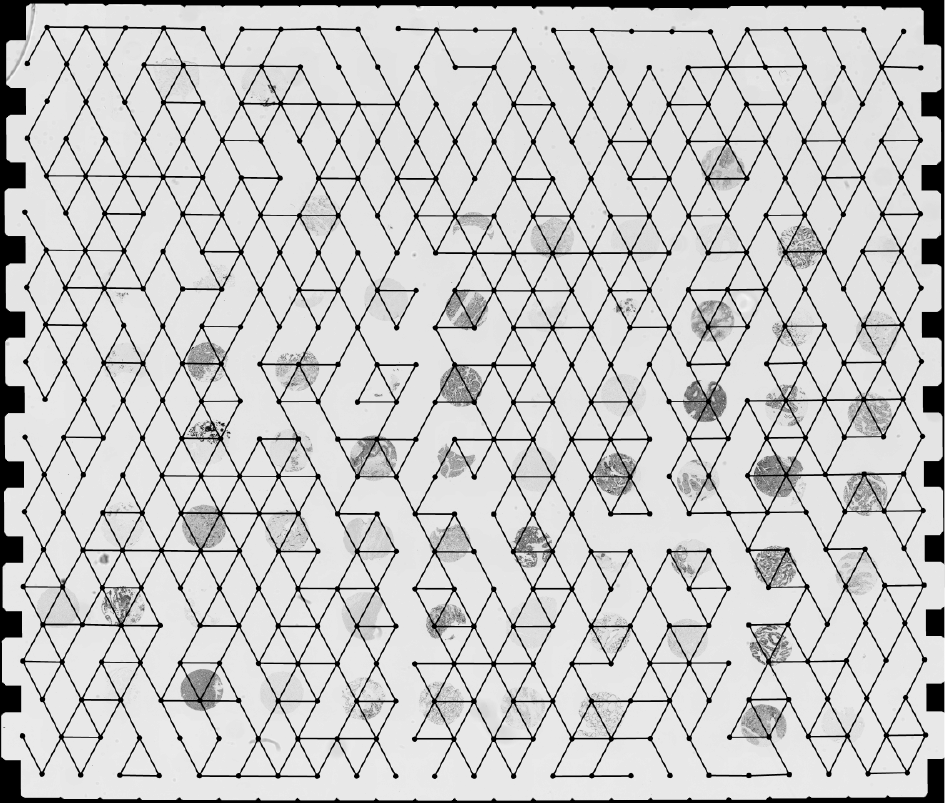}}
\caption{On the left is the final connectivity graph with the standard method, on the right with our method. Our method is able to produce a connectivity graph with just one connected component, while the standard method tuned to avoid false pairwise image registrations splits the connectivity graph into multiple connected components. With both methods, the tissue regions are stitching artifact-free, but the global placement of the tissue regions are certainly inaccurate with the standard method. With our method the placement is potentially accurate, as it is able to collect all the faint evidence from multiple uncertain pairwise registrations to produce a connectivity graph whose cycles of transformations are consistent.} 
\label{fig:tissue2stconn}
\end{figure*}

With the standard stitching method tuned to avoid false pairwise image registrations, the connectivity graph is split in multiple connected components. Therefore, the global geometry of the stitched composite image is certainly distorted, as the connected components are positioned solely based on the inaccurate estimate of the camera motion. With our method, more uncertain pairwise registrations can be included as transformation candidates, and we are able to obtain a connectivity graph with just a single connected component. This is demonstrated in Figure \ref{fig:tissue2stconn}. The resulting connectivity graph had 2202 edges, for which the dummy edge was selected for 246 edges.

\subsection{Experimental comparison of our method with the standard method}
In Table \ref{tab:summary}, we compare the root-mean-square (RMS) error in pixels, computed from the result of the minimization problem \eqref{eq:pointalign}, and the number of edges in the connectivity graph in previous experiments with the standard method and with our method.

For the standard method, where the pairwise registration corresponding to the strongest cross-correlation is selected, we consider two cases. In the first case, we do not tune the standard method to leave out uncertain low quality (LQ) registrations from the connectivity graph, but use the same threshold for the registrations as in our method. In the second case, we tune the standard method to only include certain high quality (HQ) pairwise registrations.

\begin{table}[ht]
  \centering
  \resizebox{0.98\textwidth}{!}{\begin{minipage}{\textwidth}
    \centering
  \begin{tabular}{c | c | c | c}
 &  \multicolumn{3}{c}{Method} \\ \hline 
	Experiment & Standard LQ & Standard HQ &  Ours \\ \hline
	Tissue slices & 2.7, 1558 & 2.0, 1388  & 0.45, 1472 \\ 
	Grid pattern & 35.5, 508 & 36.4, 450 & 0.55, 486 \\ 
	Tissue pattern & 9.8, 2202 & 1.3, 1238 & 0.55, 2202	
  \end{tabular}
  \end{minipage} }
  \caption{RMS error, the number of edges in the connectivity graph in standard method with low quality registrations included (LQ), without low quality registrations (HQ), and with our method.}
  \label{tab:summary}
\end{table}

By tuning pairwise registrations in the standard method, the RMS error can be decreased, but with the cost of decreasing the number of edges, i.e., the used pairwise registrations, in the connectivity graph. There is likely some error in the alignment of the corresponding image pair, which does not contribute to the RMS error.

In contrast, our method is both able to yield a small RMS error and to retain a large number of edges in the connectivity graph. This balance can be controlled by the cycle inconsistency threshold $\tau$ in the minimization problem \eqref{eq:loss}, which is reflected in the similarity of the obtained RMS error values. Increasing the threshold increases the RMS error and the number of edges, and vice versa.

\section{Conclusion} \label{seq:conclusion}
A new stitching method was presented that is robust against ambiguous pairwise image matches without dropping the ambiguous image matches entirely. Instead, multiple match candidates could be considered, and one of them was selected only if it is consistent with the rest of the connectivity graph formed by the pairwise image registration. The problem was formulated as a nonlinear minimization problem with a linear equality constraint, and an efficient minimization method was developed to solve the problem.

The method has the greatest impact when the connectivity graph has an abundance of cycles, for example in the scanning applications where each image typically has four to eight neighbors to attempt to match the image with. At minimum, the connectivity graph should have at least one cycle. This is satisfied, for example, when stitching a 360 degree panorama.

The main contribution of this work is that it makes also the second major phase, the global alignment, of the image stitching process more robust to outliers. This alleviates the need of the first phase, the pairwise registration, to be completely error free. Indeed, an overly careful pairwise registration might result in filtering out the true image registrations, which in turn would decrease the stitching quality.

\bibliographystyle{ieeetr}
\bibliography{stitching}

\end{document}